\documentclass[10pt,twocolumn,letterpaper]{article}

\usepackage[pagenumbers]{cvpr} 

\usepackage{graphicx}
\usepackage{amsmath}
\usepackage{amssymb}
\usepackage{booktabs}

\usepackage[pagebackref,breaklinks,colorlinks]{hyperref}

\usepackage[capitalize]{cleveref}
\crefname{section}{Sec.}{Secs.}
\Crefname{section}{Section}{Sections}
\Crefname{table}{Table}{Tables}
\crefname{table}{Tab.}{Tabs.}

\def\cvprPaperID{5850} 
\def\confName{CVPR}
\def\confYear{2023}

\begin{document}

\title{AttriCLIP: A Non-Incremental Learner for Incremental Knowledge Learning} 

\author{
Runqi Wang{$^{1,2}$}{$^*$}, Xiaoyue Duan{$^1$}\thanks {Co-First Author.}, Guoliang Kang{$^{1,4}$}, Jianzhuang Liu{$^{2}$},\\
Shaohui Lin{$^{3}$}, Songcen Xu{$^{2}$}, Jinhu Lv{$^{1,4}$}, Baochang Zhang{$^{1,4}$}\thanks {Corresponding Author.} \\ 
{$^1$}Beihang University \ {$^2$}Huawei Noah's Ark Lab \\ {$^3$}East China Normal University \ {$^{4}$}Zhongguancun Laboratory, Beijing China}

\maketitle
\begin{abstract}
\vspace{-3mm}
Continual learning aims to enable a model to incrementally learn knowledge from sequentially arrived data. Previous works adopt the conventional classification architecture, which consists of a feature extractor and a classifier.
The feature extractor is shared across sequentially arrived tasks or classes,
but one specific group of weights of the classifier corresponding to one new class should be incrementally expanded. Consequently, the parameters of a continual learner gradually increase. Moreover, as the classifier contains all historical arrived classes, a certain size of the memory is usually required to store rehearsal data to mitigate classifier bias and catastrophic forgetting. In this paper, we propose a non-incremental learner, named AttriCLIP, to incrementally extract knowledge of new classes or tasks. Specifically, AttriCLIP is built upon the pre-trained visual-language model CLIP. Its image encoder and text encoder are fixed to extract features from both images and text. Text consists of a category name and a fixed number of learnable parameters which are selected from our designed {attribute word bank} and serve as attributes. As we compute the visual and textual similarity for classification, AttriCLIP is a non-incremental learner. The attribute prompts, which encode the common knowledge useful for classification, can effectively mitigate the catastrophic forgetting and avoid constructing a replay memory. We evaluate our AttriCLIP and compare it with CLIP-based and previous state-of-the-art continual learning methods in realistic settings with domain-shift and long-sequence learning. The results show that our method performs favorably against previous state-of-the-arts.

\vspace{-3mm}
\end{abstract}

\vspace{-3mm}
\section{Introduction}
\label{sec:introduction}

\begin{figure*}[ht]
    \begin{center}  \includegraphics[width=0.8\linewidth]{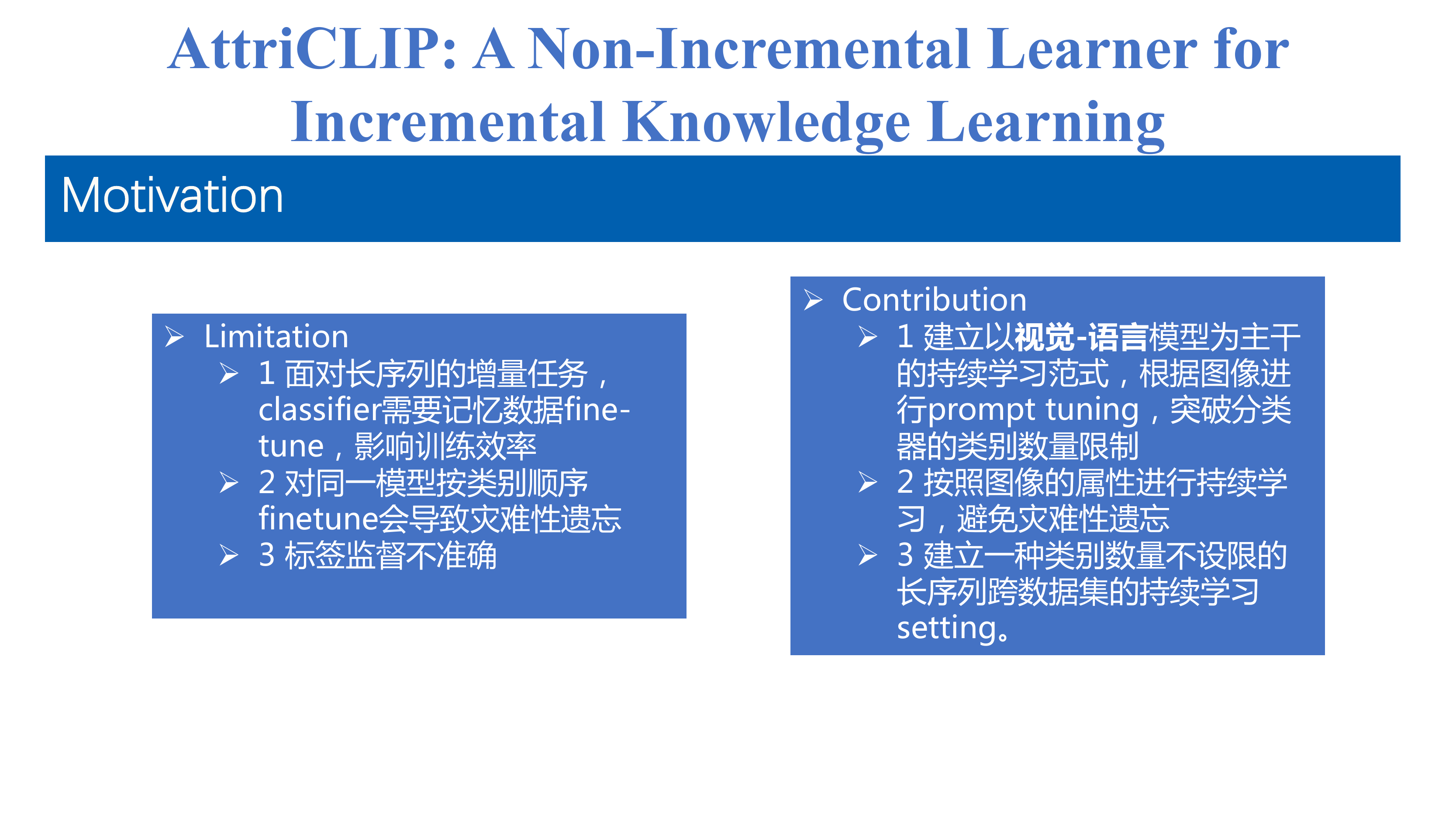}
    \vspace{-3mm}
    \caption{(a) Traditional framework for continual learning. The encoder and the classifier are trained by tasks in sequence, some of which even need extra memory data. In the framework, the model parameters of the current task are fine-tuned from the parameters trained by the previous last task and then are used for the classification of all seen tasks. The total number of categories the model can classify is fixed in the classifier. (b) Our proposed AttriCLIP for continual learning. AttriCLIP is based on CLIP, which classifies images by contrasting them with their descriptive texts. The trainable prompts are selected by the attributes of the current image from a prompt pool. The  prompts are different if the attributes of the image are different. The trained prompts are concatenated with the class name of the image, which serve as a more accurate supervised signal for image classification than labels.}
    \label{fig:motivation}
\end{center}
\vspace{-6mm}
\end{figure*}

\vspace{-2mm}
In recent years, deep neural networks have achieved remarkable progress in classification when all
the classes (or tasks) are jointly trained. 
However, in real scenarios, the tasks or classes usually sequentially arrive. 
Continual learning~\cite{li2017learning, rebuffi2017icarl,wang2022anti} aims to train a model which incrementally
expands its knowledge so as to deal with all the historical tasks or classes, behaving as if those tasks or classes are jointly trained. 
The conventional continual learning methods learn sequentially arrived tasks or classes with a shared model, as shown in Fig.~\ref{fig:motivation}(a). Such processing that fine-tunes the same model
in sequence inevitably results in subsequent values of the parameters overwriting previous ones~\cite{rajasegaran2020itaml}, which leads to catastrophic forgetting. Besides, the classification ability on historical data can be easily destroyed by current-stage learning.
In the conventional continual learning methods, a classifier on top of the feature extractor is employed to perform recognition. 
As one group of weights in the classifier is responsible for the prediction of one specific class, 
the classifier needs to be expanded sequentially to make a continual learner able to recognize novel classes.  
Moreover, extra replay data is usually required to reduce the classifier bias and the catastrophic forgetting of learned features.
It is still challenging if we expect a \textit{non-incremental learner}, \emph{i.e.}, 
the trainable parameters of the model do not incrementally increase and no replay data is needed to avoid the classifier bias and the catastrophic forgetting.

To address the above issues, this paper proposes a continual learning method named AttriCLIP , which adopts the frozen encoders of CLIP~\cite{radford2021learning}. It is a typical visual-language model that conducts image classification by 
contrasting the features of images and their descriptive texts.
In the face of increasing instances, we design a prompt tuning scheme
for continual learning. As shown in Fig.~\ref{fig:motivation}(b), there are similar attributes in images with different categories, such as ``a brown-white dog lying on the grass" and ``a brown-white cat lying on the grass".
They belong to different categories but both have ``lying on the grass" and ``brown-white" attributes, so the distance between these two images of different categories may be close in the feature space. Therefore, we selectively train different prompts based on the attributes of the images rather than the categories. In this way, there is no problem of knowledge overwriting caused by sequential training of the same model with increasing tasks.

Specifically, an {attribute word bank} is constructed as shown in Fig~\ref{fig:motivation}(b), which consists of a set of (key, prompt) pairs. {The keys represent the local features (attributes) of images and the prompts represent the descriptive words corresponding to the keys.} Several prompts are selected according to the similarities between their keys and the input image. The selected prompts are trained to capture an accurate textual description of the attributes of the image. If images with different labels have similar attributes, it is also possible to select the same prompts, \emph{i.e.}, the same prompts can be trained with images of different categories. Similarly, different prompts can be trained by the images of the same category. The trained prompts are concatenated with the class names and put into the text encoder to contrast with the image feature from the image encoder. 
This process makes our AttriCLIP distinct from all previous classifier-based framework as it serves as a non-incremental learner without the need to store replay data. 
The goal of our method is to select the existing attributes of the current image as the text description in the inference process, to classify the image. In addition, the structure of AttriCLIP can also avoid the problem of the increasing classifier parameters with the increase of the tasks and the problem of the inefficiency about memory data in the traditional continual learning methods.

The experimental setup of existing continual learning methods is idealized which divides one dataset into several tasks for continual learning. The model can set the output dimensionality of the classifier according to the total number of categories in the dataset. In practical applications, with the continuous accumulation of data, the total number of categories of samples usually cannot be obtained when the model is established. When the total number of categories exceeds the preset output dimensionality of the classifier, the model has to add parameters to classifier and requires the previous samples for fine-tuning, which greatly increases the training burden. Therefore, the continual learning approaches are required to have the ability to adapt to the categories of freely increasing data, \emph{i.e.}, the model capacity should not have a category upper limit. In order to measure such ability of continual learning models, we propose a Cross-Datasets Continual Learning (CDCL) setup, which verifies the classification performance of the model on long-sequence domain-shift tasks. 
The contributions of this paper are summarized as follows:

\begin{itemize}
\item 
\vspace{-2mm}
We establish AttriCLIP, which is a prompt tuning approach for continual learning based on CLIP. We train different prompts according to the attributes of images to avoid knowledge overwriting caused by training the same model in sequence of classes.

\item
\vspace{-2mm}
AttriCLIP contrasts the images and their descriptive texts based on the learned attributes. This approach avoids the memory data requirement for fine-tuning the classifier of increasing size.

\item
\vspace{-2mm}
In order to evaluate the performance of the model on long-sequence domain-shift tasks, we propose a Cross-Datasets Continual Learning (CDCL) experimental setup. AttriCLIP exhibits excellent performance and training efficiency on CDCL.

\end{itemize}

\vspace{-4mm}
\section{Related Work}
\label{sec:related work}
\vspace{-1.5mm}

\textbf{Continual learning.}
The existing continual learning algorithms can be mainly divided into three categories: regularization-based, architecture-based, and rehearsal-based methods.
Regularization-based methods~\cite{kirkpatrick2017overcoming, aljundi2018memory, li2017learning} alleviate catastrophic forgetting to some extent by putting constraints on important parameters related to previous tasks 
without any memory replay. 
The core of architecture-based approaches is to assign independent parameters to different tasks, which can be achieved either by expanding a network~\cite{yoon2017lifelong, rusu2016progressive, li2019learn}, or by dividing the model into sub-networks~\cite{mallya2018packnet, serra2018overcoming, wang2020learn}. However, most methods are not applicable to task-agnostic settings, where the task identity is unknown during inference. Even if the problem can be partially solved by recent methods~\cite{wortsman2020supermasks, pham2021dualnet, yan2021dynamically}, these methods are not lightweight enough.
Rehearsal-based methods store the data of previous tasks in a so-called rehearsal buffer to train with the current task. Simple yet effective, these methods achieve impressive performance on challenging settings~\cite{buzzega2020dark, cha2021co2l}. However, the performance degrades as the buffer size reduces~\cite{cha2021co2l}, and the approach to store data limits the application in privacy-sensitive scenarios~\cite{shokri2015privacy}. 

\textbf{Prompt tuning for continual learning.}
Recent continual learning works~\cite{wang2022learning, wang2022dualprompt} adopt visual prompt tuning to continual learning, which applies a small set of learnable parameters to the input, so as to provide additional instructions for the pre-trained model to be better transferred to downstream tasks~\cite{li2021align}. L2P~\cite{wang2022learning} first connects visual prompting with continual learning, and proposes to adapt the model to sequential tasks via a shared prompt pool. Inspired by the complementary learning systems, DualPrompt~\cite{wang2022dualprompt} proposed a different approach to append complementary visual prompts to the pre-trained backbone to learn task-invariant and task-specific instructions, further boosting the performance.

The prompts of L2P and DualPrompt are only attached to the image embeddings. Recent progress in vision-language models (\emph{e.g.}, CLIP~\cite{radford2021learning}) shows that language usually contains information complementary to vision. CLIP adopts a dual-stream architecture, which encodes image and text inputs into visual and textual representations in a joint embedding space. CLIP can generalize well across multiple downstream tasks. However, to the best of our knowledge, there is no work that takes
text prompts into consideration in visual continual learning. 
We propose a CLIP-based prompt tuning method, which can continuously learn long-sequence tasks effectively and efficiently based on the learned attributes of images.

\begin{figure*}[ht]
    \begin{center}  \includegraphics[width=0.8\linewidth]{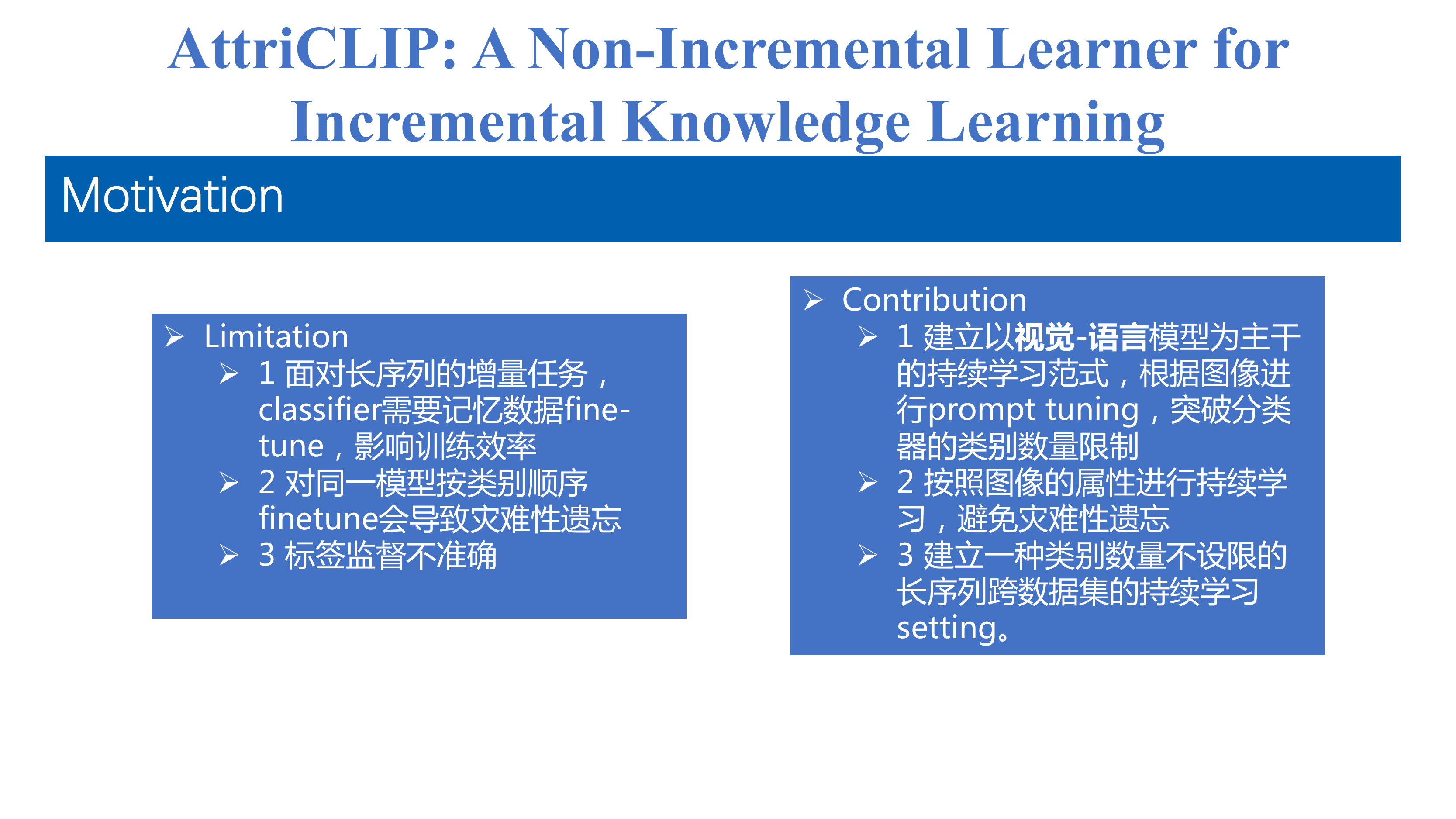}
    \vspace{-1mm}
    \caption{{Framework of AttriCLIP. The image keys $\mathbf{k}_i$ and the textual prompts $\mathbf{P}_i$ in the {attribute word bank} are trainable parameters. The blue and green boxer represent the image and text streams, respectively. The {attribute word bank} is optimized by three loss functions. $\mathcal{L}_m$ is the classification loss adopted to maximize the similarity between image feature $\mathbf{z}$ and the corresponding text features $\mathbf{w}$. $\mathcal{L}_k$ is designed to shorten the distance between the selected keys (\emph{e.g.}, $\mathbf{k}_2$ and $\mathbf{k}_n$) and the image feature $\mathbf{z}$, so that the keys learn generalizable attributes. $\mathcal{L}_p$ makes the embeddings of the prompts $g_\mathbf{\psi}(\mathbf{P}_i)$ orthogonal to increase the diversity of the prompts.}
    }
    \label{fig:framework}   
\end{center}
\vspace{-7mm}
\end{figure*}

\vspace{-1mm}
\section{Methodology}
\label{sec:methodology}
\vspace{-0.5mm}

\subsection{Preliminaries}
\label{sec3.1: preliminaries}
\vspace{-1mm}

\textbf{Continual learning formulation.} Continual learning (CL) requires a model to continuously learn new knowledge from sequential tasks without forgetting the knowledge from previous ones. Consider a sequence of tasks $\mathcal{D}\!=\!\{\mathcal{D}_1,\dots,\mathcal{D}_T\}$, where the $t$-th task $\mathcal{D}_t\!=\!\{(\mathbf{x}_i^t,y_i^t)\}_{i=1}^{n_t}$ contains $n_t$ samples $\mathbf{x}_i^t$ and the corresponding labels $y_i^t$. 
During training on task $\mathcal{D}_t$, the access to data from $\{\mathcal{D}_{1},\dots,\mathcal{D}_{t-1}\}$ is unavailable or limited.
In task-agnostic class-incremental learning,  data $\mathcal{D}_t$ of different tasks arrives in sequence $t\!=\!\{1,\dots,T\}$ without overlapping and the data arriving at different times comes from different classes. Besides, in task-agnostic setting, task identity is unknown at inference. Our AttriCLIP effectively tackles the settings above since it learns key attributes of images and allows unlimited number of output classes.

\textbf{Prompt learning based on CLIP.} CLIP~\cite{radford2021learning} consists of an image encoder $f_\mathbf{\theta}(\cdot)$ and a text encoder $g_\mathbf{\psi}(\cdot)$. Specifically, the image $\mathbf{x}\!\in\!\mathbb{R}^{H\times W\times C}$ and the text $\mathbf{t}\!\in\!\mathbb{R}^D$ are fed into $f_\mathbf{\theta}(\cdot)$ and $g_\mathbf{\psi}(\cdot)$ respectively to obtain the image embedding $\mathbf{z}\!\in\!\mathbb{R}^{D}$ and the text embedding $\mathbf{w}\!\in\!\mathbb{R}^{D}$, where $\mathbf{t}$ is the input word token. In CLIP, $\mathbf{t}$ is obtained via one of the hand-crafted prompts which have a template like ``a photo of a [CLS]", where [CLS] is the class name of the testing image. Thus, the probability of predicting the testing image $\mathbf{x}$ as the class $y_i$ can be computed as:
\vspace{-2mm}
\begin{equation}
p(y_i|\mathbf{x})=\frac{e^{\langle\mathbf{z},\mathbf{w}_{y_i}\rangle/\tau}}{\sum_{k=1}^K e^{\langle\mathbf{z},\mathbf{w}_k\rangle/\tau}},
\label{eq:img_text_sim}
\vspace{-2mm}   
\end{equation}
where $\tau$ is a temperature parameter learned by CLIP, $\langle\cdot,\cdot\rangle$ denotes the cosine similarity, $\mathbf{w}_k$ is the embedding derived from $\mathbf{t}_k$ of the $k$-th class, and $K$ is the total number of downstream dataset classes.

To bring about further improvements of CLIP’s performance on downstream tasks, prompt learning has been proposed to replace the hand-crafted prompt templates with a set of continual learnable vectors $\mathbf{P}$. The knowledge of downstream data is encoded into these vectors to instruct the model to better perform downstream tasks. Specifically, CoOp~\cite{zhou2022learning} concatenates $\mathbf{P}$ with the embedding of a class name, formulating the text description $\mathbf{t}_k(\mathbf{P})$ of the $k$-th class as:
\vspace{-1mm}
\begin{equation}
    \mathbf{t}_k(\mathbf{P}) = [\mathbf{p}]_1[\mathbf{p}]_2\dots[\mathbf{p}]_M[\mathbf{CLS}]_k,
    \label{eq:coop_prompt}
    \vspace{-1mm}
\end{equation}
where each $[\mathbf{p}]_m\!\in\!\mathbb{R}^D$, $m\!\in\!\{1,\dots,M\}$, is a learnable token of $\mathbf{P}$, and $\mathbf{P}\!\in\!\mathbb{R}^{D\times M}$ is shared among all classes. $[\mathbf{CLS}]_k$ is the text embedding of the $k$-th class name, which can also appear at the start and middle of the prompt. In this way, $\mathbf{w}_k$ in Eq.~\ref{eq:img_text_sim} is replaced by $g_\mathbf{\psi}(\mathbf{t}_k(\mathbf{P}))$, and the probability of predicting the testing image $\mathbf{x}$ as the class $y_i$ is computed as:
\vspace{-1mm}
\begin{equation}
    p(y_i|\mathbf{x})=\frac{e^{\langle\mathbf{z},g_\mathbf{\psi}(\mathbf{t}_{y_i}(\mathbf{P}))\rangle/\tau}}{\sum_{k=1}^K e^{\langle\mathbf{z},g_\mathbf{\psi}(\mathbf{t}_k(\mathbf{P}))\rangle/\tau}},
    \label{eq:coop}
\end{equation}


\subsection{{Framework of AttriCLIP}}
\label{sec3.2}
\vspace{-1mm}

In CoOp, each class embedding corresponds to only one group of prompt vectors. However, images from the same class contain diverse attributes. Encoding these diverse attributes into the same group of prompts leads to catastrophic knowledge forgetting. Besides, the encoded knowledge in the prompts of CoOp cannot interact among different classes. However, the attributes of one class may help to identify another class with similar attributes. For example, given an image of a dog lying on the grass, the attribute ``on the grass" in this image may also be found in images of other animals (\emph{e.g.}, a cat lying on the grass). We believe that prompt tuning based on the image attributes can help the prompts learn the textual descriptions of these attributes, which generalize better among tasks. 

Therefore, we propose AttriCLIP as shown in Fig.~\ref{fig:framework}, which contains an {attribute word bank} to let the image itself decide which prompts to learn based on the attributes it has. 
Only a part of the prompts that are relevant to the current image attributes are selected and trained at a time.
The {attribute word bank stores visual and textual information,} which consists of $N$ (key, prompt) pairs:
\begin{equation}
     \{\mathcal{K},\mathcal{P}\} \triangleq \{(\mathbf{k}_1, \mathbf{P}_1),\dots,(\mathbf{k}_N,\mathbf{P}_N)\},
\end{equation}
where $\{\mathcal{K},\mathcal{P}\}$ denotes the {attribute word bank}, each $\mathbf{k}_i\!\in\!\mathbb{R}^D$ has the same dimensionality as the image embedding $\mathbf{z}$, and each $\mathbf{P}_i\!=\![\mathbf{p}_i]_1\dots[\mathbf{p}_i]_M\!\in\!\mathbb{R}^{D\times M}$ is composed of $M$ learnable vectors. Denoting the set of all keys as $\mathcal{K}\!=\!\{\mathbf{k}_i\}_{i=1}^N$ and the set of all prompts as $\mathcal{P}\!=\!\{\mathbf{P}_i\}_{i=1}^N$. 
{$\mathcal{K}$ indicate the image attributes, and $\mathcal{P}$ indicate the prompt words.}
Ideally, we expect that the image itself can decide which prompts should be chosen based on the attributes it contains to guide the prediction. To this end, given an input image $\mathbf{x}_j$, we first obtain its image embedding $\mathbf{z}_j\!=\!f_\mathbf{\theta}(\mathbf{x}_j)$, where $j$ is the index of the image. Then, by scoring the match between $\mathbf{z}_j$ and $\mathbf{k}_i$ via a scoring function $\gamma$ (\emph{e.g.}, cosine distance), we select the top-$C$ keys that match $\mathbf{z}_j$ most by:
\begin{equation}
    \mathcal{K}_j=\text{Top-}C^{min}\{\gamma(\mathbf{z}_j, \mathbf{k}_{j_i})\}_{i=1}^N,
    \label{eq:choose top c}
\end{equation}
where $\text{Top-}C^{min}$ denotes the operation of choosing the top-$C$ minimal values for a set. $\mathcal{K}_j$ denotes the subset of top-$C$ keys selected from $\mathcal{K}$ specifically for the $j$-th image. We then choose the corresponding prompts that are paired with these keys, denoted as $\mathcal{P}_j\!=\!\{\mathbf{P}_{j_i}\}_{i=1}^C$, {where $\mathbf{P}_{j_i}$ is the $i$-th prompt selected specifically for $\mathbf{x}_j$.} These prompts are attached to the class name embedding of $\mathbf{x}_j$ as illustrated in Fig.~\ref{fig:framework}, and $\mathbf{t}_k(\mathbf{P})$ in Eq.~\ref{eq:coop_prompt} is then denoted as:
\begin{equation}
    \vspace{-1mm}
    \mathbf{t}_k(\mathcal{P}_j)=\operatorname{concat}(\mathbf{P}_{j_1};\dots;\mathbf{P}_{j_C};[\text{CLS}]_k),
    \vspace{-1mm}
\end{equation}
where $\operatorname{concat}(\cdot)$ denotes concatenation. Therefore, given test image $\mathbf{x}_j$ and the prompts $\mathcal{P}_j$ selected according to the attributes of $\mathbf{x}_j$, the probability of predicting the image as class $y_i$ is finally computed as:
\begin{equation}
\textbf{}
    p(y_i|\mathbf{x}_j)=\frac{e^{\langle\mathbf{z},g_\mathbf{\psi}(\mathbf{t}_{y_i}(\mathcal{P}_j))\rangle/\tau}}{\sum_{k=1}^K e^{\langle\mathbf{z},g_\mathbf{\psi}(\mathbf{t}_k(\mathcal{P}_j))\rangle/\tau}}.
    \textbf{}
    \label{eq:our method}
\end{equation}

From a high-level perspective, the proposed {attribute word bank} serves as a bridge between the output of the image encoder and the input of the text encoder. 
The keys are optimized to be close to the matched image embeddings, which contain rich high-level information, \emph{i.e.}, image attributes. 
The prompts are optimized to include textual information related to the corresponding image attributes, so as to better guide the model predictions along with the class name embeddings. 
Since the proposed {attribute word bank} connects the image stream and the text stream, our prompts serve more as the textual descriptions of the image attributes compared with DualPrompt~\cite{wang2022dualprompt}. 
In addition, since the generalizable attributes are learned, the memory is no longer needed to fine-tune the classifier based on previous tasks, which makes AttriCLIP more efficient under the long sequence setting. 


\subsection{{Optimization Objective of AttriCLIP}}
\label{sec3.3}

Based on Eq.~\ref{eq:our method}, the image classification loss is formulated as:
\begin{equation}
    \mathcal{L}_m=\mathbb{E}[-\text{log}\frac{e^{\langle\mathbf{z},g_\mathbf{\psi}(\mathbf{t}_{y_i}(\mathcal{P}_j))\rangle/\tau}}{\sum_{k=1}^K e^{\langle\mathbf{z},g_\mathbf{\psi}(\mathbf{t}_k(\mathcal{P}_j))\rangle/\tau}}].
    \label{eq:Lm}
\end{equation}
In addition to $\mathcal{L}_m$, a matching loss is needed to pull the matched top-$C$ keys $\mathcal{K}_j$ closer to the image embedding $\mathbf{z}_j$, so that the keys learn rich attributes from the samples. The matching loss adopted to optimize the keys is defined as:
{\begin{equation}
    \vspace{-1mm}
    \mathcal{L}_k=\sum_{i=1}^C\gamma(\mathbf{z}_j, \mathbf{k}_{j_i}).
    \label{eq:Lk}
    \vspace{-1mm}
\end{equation}}
We test three distance functions (\emph{i.e.}, cosine distance~\cite{sanh2019distilbert}, mean square error (MSE)~\cite{marmolin1986subjective} and triplet loss~\cite{dong2018triplet}) for $\gamma$, and find that the cosine distance works the best (see Sec.~\ref{sec:4.4}). Finally, in order to make the learned prompts more semantically diverse, we adopt a third loss to orthogonalize
the embeddings of different prompts to increase the diversity of the prompts:
\begin{equation}
\mathcal{L}_p=\frac{1}{N(N-1)}\sum_{i=1}^{N}\sum_{j=i+1}^N|\langle g_\mathbf{\psi}(\mathbf{P}_i), g_\mathbf{\psi}(\mathbf{P}_j)\rangle|,
\label{eq:Lp}
\end{equation}
where $\langle\cdot,\cdot\rangle$ denotes the cosine similarity. In this way, the overall optimization objective is defined as:
\begin{equation}    \mathcal{L}=\mathcal{L}_m+\lambda_k\mathcal{L}_k+\lambda_p\mathcal{L}_p,
\label{eq:total loss}
\end{equation}
where $\lambda_k$ and $\lambda_p$ are balance factors. The keys are optimized by $\mathcal{L}_k$, and the prompts by $\mathcal{L}_m$ and $\mathcal{L}_p$.

\begin{table*}[t]
\centering
\caption{{Average accuracy~\cite{lopez2017gradient} of different continual learning methods on CIFAR100~\cite{krizhevsky2009learning}. The accuracy of Task $t,t\!\in\!\{1,2,\dots,10\}$ reported here is the test accuracy averaged over all the previous tasks (\emph{i.e.}, Tasks $1,2,\dots,t$).}
}
\vspace{-1.5mm}
\resizebox{1.8\columnwidth}{!}{
\begin{tabular}{cccccccccccc}
\toprule[1pt]
Method & Memory & Task 1 & Task 2 & Task 3 & Task 4 & Task 5 & Task 6 & Task 7 & Task 8 & Task9 & Task10 \\ \midrule[0.3pt]
LwF            & 0      & 89.3   & 70.1   & 54.3   & 45.8   & 39.8   & 36.1   & 31.7   & 28.9   & 24.4  & 23.9   \\
iCaRL          & 2000      & 88.7   & 78.1   & 72.4   & 67.2   & 63.7   & 60.2   & 56.4   & 54.4   & 51.9  & 49.5   \\
iTAML          & 2000   & 89.2   & 89.0   & 87.3   & 86.2   & 84.3   & 82.1   & 80.7   & 79.1   & 78.4  & 77.8   \\
{ARI}  & 2000    & {88.6} & {86.9} & {85.8} & {84.6} & {83.1} & {81.8} & {81.6} & {81.0} & {80.2} & {80.9}   \\
CoOp           & 1000   & 95.8   & 90.7   & 85.2   & 83.4   & 80.8   & 75.8   & 74.7   & 71.7   & 71.3  & 67.6   \\
Continual-CLIP & 0      & 96.7   & 92.2   & 86.0   & 80.4   & 77.5   & 75.8   & 73.0   & 71.4   & 69.8  & 66.7   \\
\midrule[0.3pt]
\textbf{AttriCLIP}          & 0      & \textbf{97.8} & \textbf{93.7} & \textbf{91.0} & \textbf{87.5} & \textbf{84.7} & \textbf{82.5} & \textbf{81.3} & \textbf{80.5} & \textbf{80.0} & \textbf{79.7}   \\ \midrule[0.3pt]
Upper-bound    &    -    &    -    &   -     &   -     &    -    &    -    &   -     &     -   &  -      &   -    & \textbf{86.3}   \\ \bottomrule[1pt]
\end{tabular}}
\vspace{-1.5mm}
\label{tab:cifairand}
\end{table*}

\vspace{-1mm}
\section{Experiments}
\label{sec:experiments}
\vspace{-1mm}
{In this section, the implementation details are first described. We then compare the proposed AttriCLIP with other methods in the conventional class-incremental task-agnostic setting, and the proposed Cross-Datasets Continual Learning (CDCL) setting. Finally, we perform ablation studies to evaluate the effect of different components of AttriCLIP.} We implement our model using the MindSpore Lite tool~\cite{mindspore}.
\vspace{-1mm}
\subsection{Implementation Details}
\label{sec:4.1}
\vspace{-1mm}
\textbf{Datasets.}
The experiments are conducted on CIFAR-100~\cite{krizhevsky2009learning} and ImageNet100~\cite{deng2009imagenet}.
{CIFAR100} consists of 60k images with a size of 32$\times$32 from 100 classes, which are split into 10 tasks with 10 classes in each task. Each class consists of 500 training and 100 testing samples.
{ImageNet100}, as a subset of ILSVRC2012~\cite{krizhevsky2017imagenet}, contains samples sized 224$\times$224 from 100 classes.
Each class consists of about 1,300 training and 50 test samples. We split ImageNet100 into 10 tasks with 10 classes in each task. More details of ImageNet100 are provided in the supplementary.

\begin{table*}[ht]
\centering
\caption{{Average accuracy~\cite{lopez2017gradient} of different continual learning methods on ImageNet100~\cite{deng2009imagenet}. The accuracy of Task $t,t\!\in\!\{1,2,\dots,10\}$ reported here is the test accuracy averaged over all the previous tasks (\emph{i.e.}, Tasks $1,2,\dots,t$).}
}
\vspace{-1.5mm}
\resizebox{1.8\columnwidth}{!}{
\begin{tabular}{cccccccccccc}
\toprule[1pt]
    Method        & Memory & Task 1        & Task 2        & Task 3        & Task 4        & Task 5        & Task 6        & Task 7        & Task 8        & Task9         & Task10        \\ \midrule[0.3pt]
iCaRL       & 2000      & 82.1          & 80.6          & 75.5          & 70.1          & 68.1          & 65.8          & 62.5          & 61.3          & 60.7          & 59.5          \\
DER       & 2000      & 81.7          & 80.6          & 76.0          & 72.1          & 74.4          & 71.8          & 70.5          & 68.3          & 67.3          & 66.7          \\
ARI   & 2000   & 87.6          & 85.4          & 83.1          & 82.6          & 80.4          & 80.8          & 80.5          & 80.1          & 79.6          & 79.3        \\
CoOp        & 1000   & 89.2          & 83.2          & 76.7         & 79.8         & 79.9         & 82.34         & 79.7         & 80.1          & 80.3          & 79.3         \\
Continual-CLIP        & 0   &  93.3         &87.6           & 83.1        &  81.7        &   80.5       &   80.2       & 79.3         &    78.5       &  76.9         &  75.4        \\ \midrule[0.3pt]
\textbf{AttriCLIP}      & 0      & \textbf{95.4} & \textbf{89.4} & \textbf{88.9} & \textbf{88.5} & \textbf{85.6} & \textbf{87.5} & \textbf{86.9} & \textbf{86.6} & \textbf{87.3} & \textbf{81.4} \\ \midrule[0.3pt]
Upper-bound &    -    &        -       &       -        &        -       &       -        &      -         &    -           &      -         &           -    &           -    & \textbf{91.4}          \\ \bottomrule[1pt]
\end{tabular}}
\vspace{-3mm}
\label{tab:imarand}
\end{table*}

\textbf{Baselines.} We compare the proposed AttriCLIP with existing CLIP-based methods (CoOp~\cite{zhou2022learning} and continual-CLIP~\cite{thengane2022clip}), prompt-based methods (DualPrompt~\cite{wang2022dualprompt}) and typical continual learning methods (LwF~\cite{li2017learning}, iCaRL~\cite{rebuffi2017icarl}, DER~\cite{yan2021dynamically}, iTAML~\cite{rajasegaran2020itaml} and ARI~\cite{wang2022anti}). 
The prompts of CoOp are trained in a sequence of tasks and store partial data from previous tasks in the memory for the prompts fine-tuning on subsequent tasks. The continual-CLIP evaluates a frozen pre-trained CLIP model in continual learning settings. 
iCaRL is a classic method of continual learning with memory data. ARI is the current state-of-the-art for non-prompt-based continual learning methods. We adopt ViT-L-14~\cite{he2016deep} as the backbone for CoOp, continual-CLIP, DualPrompt and our AttriCLIP, and adopt ResNet~\cite{dosovitskiy2020image} for other methods.
{All the methods are evaluated under the task-agnostic setting, and our proposed AttriCLIP does not need any memory, which makes the setting more practical and challenging.}

\textbf{Training details.} 
{We train the model for 10 epochs on each incremental task for all datasets.}
SGD is adopted as the optimizer with the initial learning rate set to 0.001 and following a cosine decay schedule. The weight decay is 0, the batch size is 32, and the loss weights $\lambda_k$ and $\lambda_p$ are $0.7$ and $0.3$ respectively. The prompt length $M\!=\!12$, the number of attributes in the bank $N\!=\!10$ and the number of selected attributes $C\!=\!3$. 
{The average results over 3 runs are reported for all methods.}

\vspace{-1mm}
\subsection{Results of Class-Incremental Learning}
\label{sec:4.2}
\vspace{-1mm}

{In this section, we take the average accuracy~\cite{lopez2017gradient}, as the metric to measure the performance. The buffer size of data from previous tasks is denoted as Memory. We compare the proposed AttriCLIP with the prior arts on CIFAR100, and the results are reported in Table~\ref{tab:cifairand}. From the results, we see that AttriCLIP achieves the best average accuracy compared with the recent state-of-the art methods such as ARI and Continual-CLIP. Besides, compared with previous CLIP-based methods (\emph{i.e.}, CoOp and Continual-CLIP), AttriCLIP outperforms them by a large margin. Specifically, AttriCLIP outperforms CoOp by 13.8\% without the need for any memory. It also outperforms Continual-CLIP by 14.7\%. Note that ``Upper-bound’’ in Table~\ref{tab:cifairand} denotes the standard supervised training on the data from all tasks, which is usually regarded as the upper bound of the performance one method can achieve. Compared with the upper-bound, the accuracy of AttriCLIP drops only 4.9\%, demonstrating the effectiveness of learning image attributes for mitigating catastrophic forgetting.}

\begin{table}[t]
\centering
\caption{{Accuracy of different methods on CIFAR100. The models are either trained from scratch on CIFAR100 (CIFAR100), or fine-tuned on CIFAR100 after being continually trained from scratch on ImageNet100 (CIFAR100-I2C).}
}
\resizebox{1\columnwidth}{!}{
\begin{tabular}{ccccc}
\toprule[1pt]
Method      & Memory & CIFAR100 & CIFAR100-I2C & FT           \\ \midrule[0.3pt]
\textit{iCaRL}-1       & 2000      & 49.5              & 49.7                   & +0.2          \\
\textit{iCaRL}-2       & 2000      & 49.1              & 46.5                   & -2.6          \\
\textit{CoOp}-1        & 1000   & 67.6              &   61.1                  & -6.5         \\
\textit{CoOp}-2        & 1000   &  67.6                 &  59.0                     &     -8.6          \\
\textit{ARI}-1         & 2000   & 80.9              &    74.5                 & -6.4          \\
\textit{ARI}-2         & 2000   &  79.7               &     59.9                  &    -19.8           \\
Continual-CLIP         & 0   & 66.7                &   66.7                   &    0          \\
\textit{DualPrompt}-1 & 0      &    \textbf{86.5}               & 80.7                     &   -5.8         \\
\textit{DualPrompt}-2 & 0      &    {84.1}               &     74.7                     &     -9.4            \\\midrule[0.3pt]
\textbf{AttriCLIP}      & 0      & 81.4     & \textbf{82.3}          & \textbf{+0.9} \\ \bottomrule[1pt]
\end{tabular}}
\vspace{-5mm}
\label{tab:cifarima}
\end{table}

{We also compare our method with previous arts on ImageNet100, and report the results in Table~\ref{tab:imarand}. AttriCLIP still outperforms other memory-based methods without any memory needed. For example, AttriCLIP outperforms ARI, which was the state-of-the-art, by 4.0\%. Compared with CLIP-based models, our method again outperforms CoOp and Continual-CLIP by 4.0\% and 7.9\% respectively. The accuracy of AttriCLIP decreases by 8.1\% compared with the upper-bound. The results suggest that the effectiveness of our method over different datasets.}

\vspace{-1mm}
\subsection{Results of Cross-Datasets Continual Learning}
\label{sec:4.3}
\vspace{-1mm}
{To simulate the practical setting where the model continuously learns long sequence tasks, we propose a new setting for evaluation, \emph{i.e.}, Cross-Datasets Continual Learning (CDCL).}
{In CDCL, the model is continually trained on several datasets one by one in a sequential manner, and then the accuracy on each dataset is evaluated.}

\textbf{Learning attributes helps the model to generalize better to a new dataset.}
{For comparison, we train two benchmarks for each method in Table~\ref{tab:cifarima}: (1) CIFAR100 benchmark, where the model is trained from scratch on CIFAR100 under the same setting (10 tasks) as in Table~\ref{tab:cifairand}, then evaluated on CIFAR100. (2) CIFAR100-I2C benchmark, where the model is first trained from scratch on ImageNet100 under the same setting (10 tasks) as in Table~\ref{tab:imarand}, then fine-tuned on CIFAR100 under the same setting (10 tasks) as in Table~\ref{tab:cifairand}, and finally evaluated on CIFAR100. We define FT (Forward Transfer) in Table~\ref{tab:cifarima} as the accuracy on CIFAR100-I2C minus the accuracy on CIFAR100, which indicates the model’s superior ability to transfer knowledge from the previous dataset to the new dataset. }


{Conventional continual learning methods adopt the typical classification architecture, which includes a classifier with a preset output dimensionality. However, in practical settings, the number of sequentially arriving classes is unlimited. Therefore, the classifier needs to be incrementally expanded to learn new tasks or classes. In this experiment, we use two schemes to expand the classifier for each classifier-based continual learning method: (1) \textit{Method}-1, \emph{i.e.}, increase the number of classifiers as the number of arriving classes increases; (2) \textit{Method}-2, \emph{i.e.}, directly increase the output dimensionality of a classifier so that it can handle more classes of data.}

{In our experiments, for \textit{Method}-1, after training a classifier (with the output dimensionality as 100) on CIFAR100, we incrementally train another classifier (with the output dimensionality also as 100) on ImageNet100. These two classifiers share the same feature extractor. Besides, when training the classifier for ImageNet100, the data from CIFAR100 is unavailable in the memory bank for memory-based methods. On the other hand, for \textit{Method}-2, we directly train one classifier but double its output dimensionality (\emph{i.e.}, 200), so that it can jointly predict data from both CIFAR100 and ImageNet100. Besides, when training on ImageNet100, the data from CIFAR100 is available in the memory bank for memory-based methods. Note that the CoOp model does not have a classifier, so the only difference between \textit{CoOp}-1 and \textit{CoOp}-2 is whether the data from the previous dataset is available in the memory bank.}

{We report the experimental results in Table~\ref{tab:cifarima}. The results demonstrate that AttriCLIP is outperformed by \textit{DualPrompt}-1 and \textit{DualPrompt}-2 on CIFAR100. It is to be noted that the encoder of \textit{DualPrompt}-1 and \textit{DualPrompt}-2 is pre-trained on ImageNet-21k~\cite{dosovitskiy2020image}. AttriCLIP still significantly exceeds the remaining methods. Specifically, AttriCLIP outperforms \textit{CoOp}-1 and \textit{CoOp}-2 by 13.8\%. Moreover, we find that among all methods, AttriCLIP is the only one that effectively transfers the knowledge learned from ImageNet100 to improve the performance on CIFAR100 (\emph{i.e.}, FT$>$0), and exceeds all other methods under the CIFAR100-I2C setting. This indicates that AttriCLIP effectively learns crucial image attributes into the prompts in the previous dataset, which helps it to generalize better to a new dataset.}


\begin{table}[t]
\centering
\vspace{-3mm}
\caption{{Accuracy of different methods on ImageNet100. The models are either trained from scratch on ImageNet100 (ImageNet100), or fine-tuned on CIFAR100 after being continually trained from scratch on ImageNet100 (ImageNet100-I2C).}
}
\resizebox{1\columnwidth}{!}{
\begin{tabular}{ccccc}
\toprule[1pt]
Method                                              &Memory          & ImageNet100 & ImageNet100-I2C& BT \\ \midrule[0.3pt]
\textit{iCaRL}-1                                                      &2000   & 59.5        & 34.5 &  {-25.0}   
\\
\textit{iCaRL}-2                                                     &2000     & 58.7        & 50.9 &  -7.8                                                                    \\
\textit{CoOp}-1                                                    &1000       & 79.3        & 57.6 &   -21.7                                                                     \\
\textit{CoOp}-2     &1000   & 79.3       & 75.9 &   -3.4                                                                   \\
\textit{ARI}-1      &2000                                                      & 79.3        & 51.2 & -28.1                                                                    \\
\textit{ARI}-2   &2000     & 77.9        & 61.8 &  -16.1                                                                  \\
Continual-CLIP         & 0   &   75.4              &      75.4                &     0         \\
\textit{DualPrompt}-1                                             &0    &    \textbf{85.4}            &  63.6    &  -21.8                                                                         \\
\textit{DualPrompt}-2                                             &0       &    81.9           &  77.8    &   -4.1                                                                \\ \midrule[0.3pt]
\textbf{AttriCLIP}                                                 &0         & 83.3        & \textbf{90.3} & \textbf{+7.0}                                                               \\\bottomrule[1pt]
\end{tabular}}
\vspace{-1mm}
\label{tab:inference}
\end{table}

\begin{table}[t]
\centering
\vspace{-3mm}
\caption{Comparison among different methods on ImageNet100 + CIFAR100 where each model is continually trained on ImageNet100 and CIFAR100 in sequence.}
\resizebox{0.65\columnwidth}{!}{
\begin{tabular}{ccc}
\toprule[1pt]
Method                                                        & Memory &\begin{tabular}[c]{@{}c@{}}CIFAR100+\\ ImageNet100\end{tabular} \\ \midrule[0.3pt]
\textit{iCaRL}-1                                                         & 2000          & 30.7                                                                  \\
\textit{iCaRL}-2                                                         & 2000          & 37.6                                                                  \\
\textit{CoOp}-1                                                          & 1000            & 46.6                                                                  \\
\textit{CoOp}-2 & 1000        & 55.4                                                                  \\
\textit{ARI}-1                                                           & 2000             & 32.5                                                                  \\
\textit{ARI}-2  & 2000           & 57.3                                                                  \\
Continual-CLIP         & 0         &  54.9            \\
\textit{DualPrompt}-1                                                &    0                 &      35.4                                                                    \\
\textit{DualPrompt}-2                                                  &    0              &     67.1                                                              \\ \midrule[0.3pt]
\textbf{AttriCLIP}                                                        & 0          & \textbf{78.3}                                                                  \\\bottomrule[1pt]
\end{tabular}}
\vspace{-5mm}
\label{tab:bothdata}
\end{table}

\begin{figure*}[t]
    \centering
    \includegraphics[width=0.8\linewidth]{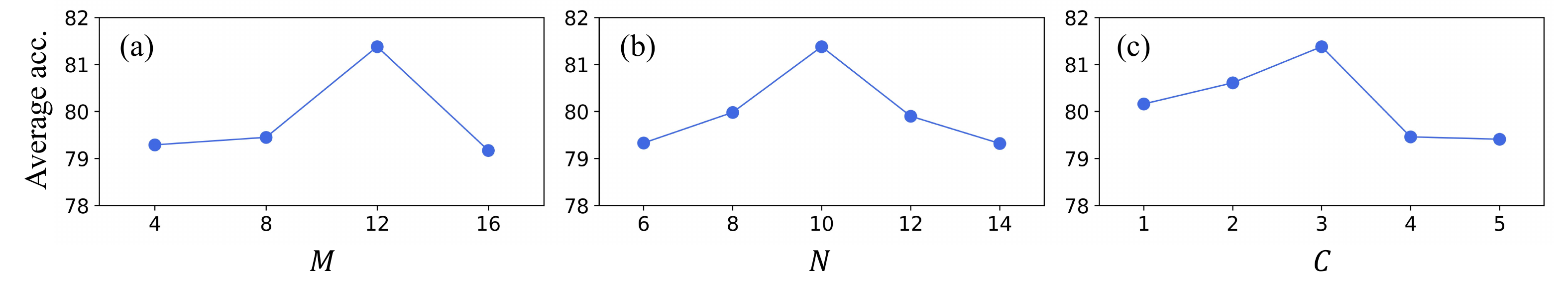}
    \vspace{-3mm}
    \caption{Ablation study of (a) the prompt length $M$, (b) the {bank} size $N$, and (c) the number of selected keys $C$ on CIFAR100.}
    \vspace{-3mm}
    \label{fig:ablation_567}
\end{figure*}

\textbf{Learning attributes helps the model NOT to forget the previous dataset.}
{In Table~\ref{tab:inference}, we test the accuracy of different continual learning methods on ImageNet100 also under two settings: (1) ImageNet100 benchmark, which is the same setting as in Table~\ref{tab:imarand}. (2) ImageNet100-I2C benchmark, which is the same as CIFAR100-I2C in training, but evaluating on ImageNet100. We define BT (Backward Transfer) as the accuracy on ImageNet100-I2C minus the accuracy on ImageNet100. The smaller the value of BT, the more knowledge from the previous dataset is forgotten after the model is trained on the new one. When BT$>$0, the model effectively transfers knowledge from the new dataset to improve the recognition performance on the previous dataset.}


{As shown in Table~\ref{tab:inference}, AttriCLIP is the only method which does not forget the knowledge from the previous dataset, and even improves the performance on the previous dataset (BT=+7.0\%). This demonstrates that our method effectively learns generalizable attributes from the new dataset, which can help the model NOT to forget, or even consolidate previously learned knowledge.}

{In addition, \textit{Method}s-1 in Table~\ref{tab:inference} forget previous knowledge more seriously than \textit{Method}-2, which indicates that replaying the data from the previous dataset or training a classifier with large output dimensionality may help to mitigate catastrophic forgetting cross datasets. However, the output dimensionality of classifier cannot be expanded endlessly, and previously trained data is often unavailable in practical settings. This again highlights the advantage of our method.}


\textbf{Learning attributes helps the model evaluate on cross datasets.}
{In Table~\ref{tab:bothdata}, we train the models following the same setting as in Table~\ref{tab:cifarima}, and finally evaluate their performances on both datasets. For a classifier-based \textit{Method}-1, given one testing image, each of its two classifiers outputs a prediction vector of 100 dimensions.
We simply choose the class with the highest score in these two output vectors as the prediction result. According to Table~\ref{tab:bothdata}, AttriCLIP achieves the highest accuracy (78.3\%) without the need for any memory, demonstrating the effectiveness of our method in the proposed CDCL setting.}


\vspace{-1mm}
\subsection{Ablations}
\label{sec:4.4}
\vspace{-1mm}
{We conduct ablation studies on CIFAR100 following the same setting as in Table~\ref{tab:cifairand}. The average accuracy over 10 tasks are reported.}


\begin{table}[t]
\centering
\caption{Comparison of different loss functions for $\mathcal{L}_k$ on CIFAR100.}
\vspace{-2mm}
\setlength{\tabcolsep}{1.2mm}{
\resizebox{0.8\columnwidth}{!}{\begin{tabular}{cccc}
\toprule[1pt]
     Loss function        & Triplet   loss & Cosine loss    & MSE loss \\ \midrule[0.3pt]
Average acc. & 80.22          & \textbf{81.38} & 80.81    \\ \bottomrule[1pt]
\end{tabular}}}
\vspace{-1mm}
\label{tab:L_k}
\end{table}

\begin{table}[t]
\centering
\caption{Average acc. of different loss weights $\lambda_k$ on CIFAR100.}
\vspace{-2mm}
\setlength{\tabcolsep}{1.8mm}{
\resizebox{0.8\columnwidth}{!}{\begin{tabular}{cccccc}
\toprule[1pt]
       $\lambda_k$      & 0.1   & 0.3   & 0.5   & 0.7            & 0.9   \\ \midrule[0.3pt]
Average acc. & 80.28 & 80.30 & 81.11 & \textbf{81.38} & 80.86 \\ \bottomrule[1pt]
\end{tabular}}}
\vspace{-2mm}
\label{tab:lamk}
\end{table}


\textbf{Loss function $L_k$.}
{We adopt three loss functions (\emph{i.e.}, triplet loss, cosine distance loss, and MSE loss) for $\gamma$ in Eq.~\ref{eq:Lk}. According to Table~\ref{tab:L_k}, the best result is obtained with cosine distance loss adopted for $\mathcal{L}_k$. We also vary the weight $\lambda_k$ of $\mathcal{L}_k$ in Eq.~\ref{eq:total loss}. The result in Table~\ref{tab:lamk} shows that the best performance (81.38\%) is achieved with $\lambda_k\!=\!0.7$.}



\textbf{Loss function $L_p$.}
{We report the results with different loss weights $\lambda_p$ of $\mathcal{L}_p$ in Table~\ref{tab:lamp}. Introducing $\mathcal{L}_p$ significantly improves the performance by increasing the diversity of prompts. The best result is obtained with $\lambda_p\!=\!0.3$.}



\begin{table}[t]
\centering
\caption{Average acc. of different loss weights $\lambda_p$ on CIFAR100.}
\vspace{-1mm}
\setlength{\tabcolsep}{1.1mm}{
\resizebox{0.8\columnwidth}{!}{\begin{tabular}{ccccccc}
\toprule[1pt]
       $\lambda_p$      & 0     & 0.1   & 0.3   & 0.5   & 0.7   & 0.9   \\ \midrule[0.3pt]
Average acc. & 78.10 & 79.23 & \textbf{81.38} & 81.28 & 81.17 & 80.88 \\ \bottomrule[1pt]
\end{tabular}}}
\vspace{-1mm}
\label{tab:lamp}
\end{table}

\textbf{The length of prompts $M$.}
{The result in Fig.~\ref{fig:ablation_567}(a) shows that the model achieves the best performance when $M\!=\!12$. When the prompt length is too long, both the training efficiency and the computation budget will be increased. }




\textbf{Bank size $N$.}
{By varying the number of (key, prompt) pairs in the {attribute word bank}, we find in Fig.~\ref{fig:ablation_567}(b) that the model achieves the best performance when $N\!=\!10$.}


\textbf{The number $C$ of attributes selected.}
{We test the effect of different values of $C$ in Eq.~\ref{eq:choose top c}, and find in Fig.~\ref{fig:ablation_567}(c) that choosing too many keys and prompts to train at the same time affects the model performance. When $C\!=\!3$, the model obtains optimal performance.}



\begin{figure}[t]
    \centering
    \includegraphics[width=0.65\linewidth]{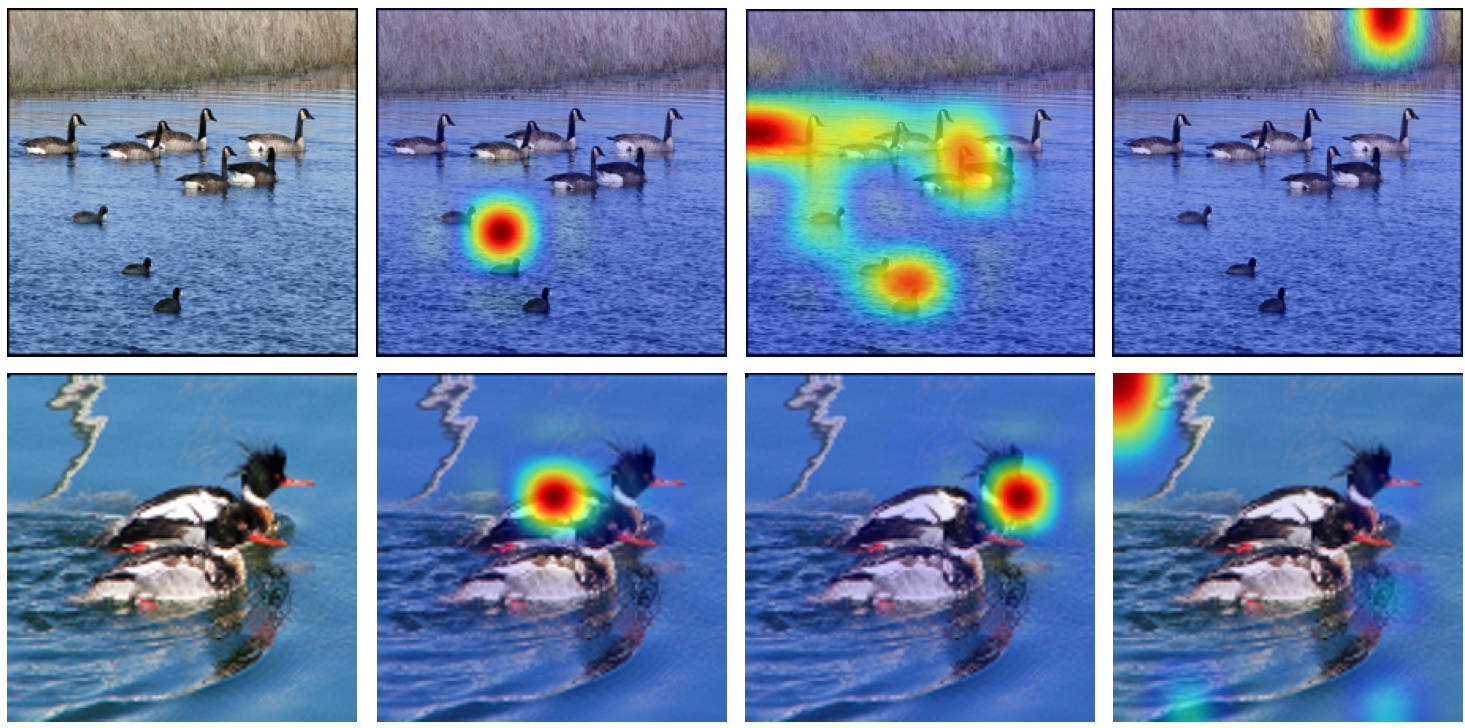}
    \caption{{Visualization of the selected prompts of the same image using Grad-CAM~\cite{selvaraju2017grad}.}
    }
    \label{fig:ima}
    \vspace{-4mm}
\end{figure}

\begin{figure}[t]
    \centering
    \includegraphics[width=0.7\linewidth]{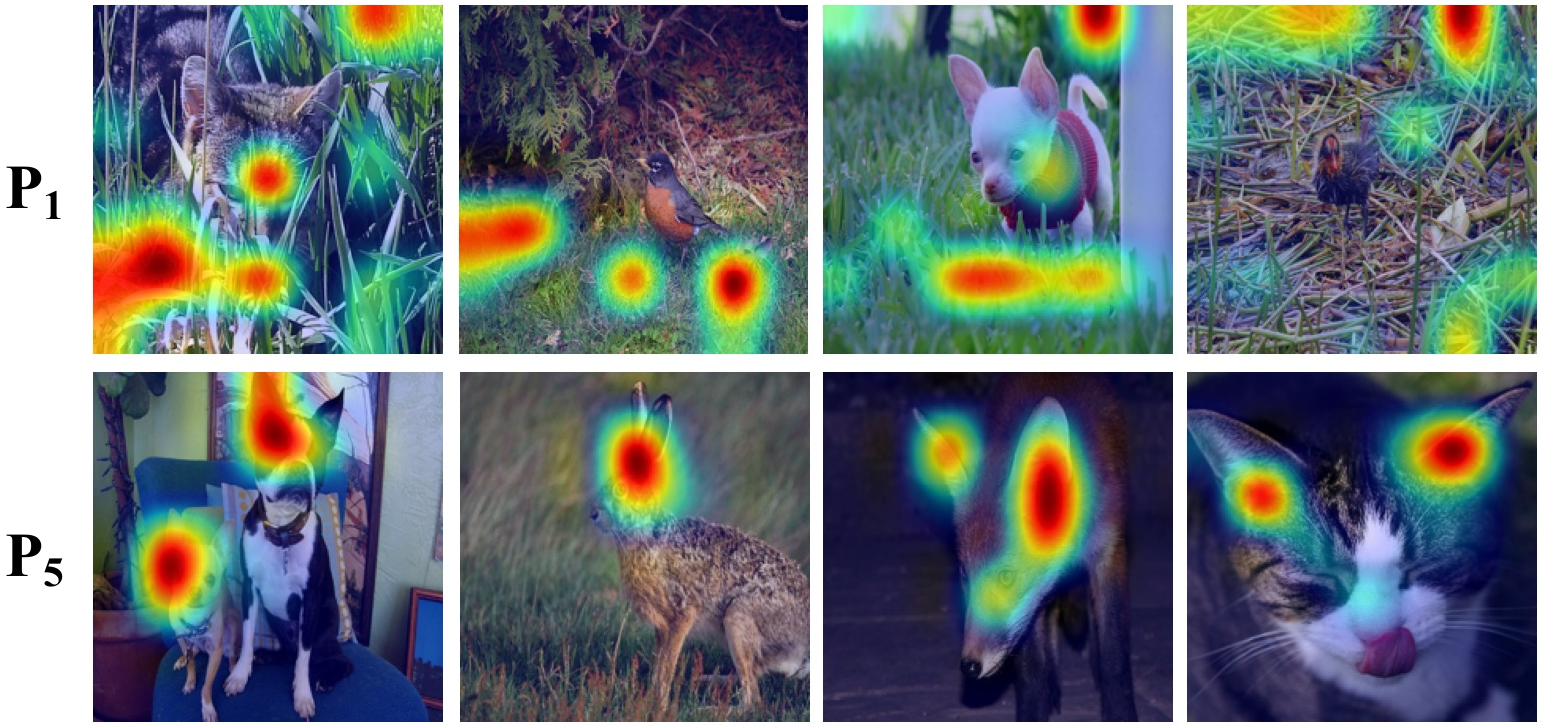}
    \caption{{Visualization of the same prompts on different images using Grad-CAM~\cite{selvaraju2017grad}.}
    }
    \vspace{-6mm}
    \label{fig:prompt}
\end{figure}

\textbf{Visualization of prompts.}
{To verify that different prompts do reflect different image attributes, we visualize the image contents corresponding to different prompts using Grad-CAM~\cite{selvaraju2017grad}. Specifically, given test image, several prompts are first selected based on the image attributes. Each selected prompt $\mathbf{P}_i$ is passed through the text encoder to obtain the prompt embedding $g_\mathbf{\psi}(\mathbf{P}_i)$. The prompt embedding and the image feature $\mathbf{z}$ are then used to calculate $\mathcal{L}_m$, which is adopted to highlight the corresponding image contents using Grad-CAM. }

{In Fig.~\ref{fig:ima}, the image contents in different columns correspond to different prompts. From Fig.~\ref{fig:ima}, it can be seen that for the same image, different prompts do reflect different regions in the image, demonstrating the diversity of the learned prompts. }


{To verify whether the learned prompts do reflect image attributes with high-level semantics, we visualize the content of the same prompts (\emph{e.g.}, $\mathbf{P}_1$ and $\mathbf{P}_5$) on different images in Fig.~\ref{fig:prompt}. It can be seen that $\mathbf{P}_1$ mainly captures the background of the images (\emph{e.g.}, the grass), while $\mathbf{P}_5$ focuses more on the foreground (\emph{e.g.}, the ears of the animals). This demonstrates that the prompts effectively learn key attributes which can generalize across images, thus improving the performance in continual learning.}


\vspace{-1mm}
\section{Conclusion}
\label{sec:conclusion}
\vspace{-1mm}
We propose a novel continual learning method, named AttriCLIP, which can incrementally learn knowledge without incrementally increasing model parameters or constructing extra memory to store replay data. 
Our framework is based on the pretrained visual-language model CLIP. We fix both the image and the text encoders, only updating the text prompts to adapt to sequentially arrived tasks or classes.
We design a module named {attribute word bank to store attributes of images and their descriptive words.}
Experiments show that our method performs favourably against vanilla CLIP, typical prompt learning methods and previous state-of-the-arts, especially in the long-sequence and cross-domain settings.
We believe our work paves the way for more practical continual learning, where we need to consider the incremental knowledge in a long task sequence or face the common domain shift.
\vspace{-1mm}
\section*{Acknowledgements}
\vspace{-1mm}
This work was supported by National Natural Science Foundation of China under Grant 62076016, 62141604 and 62102151, Beijing Natural Science Foundation L223024 and Shanghai Sailing Program (21YF1411200). We gratefully acknowledge the support of MindSpore \cite{mindspore}, CANN (Compute Architecture for Neural Networks) and Ascend AI Processor used for this research. 
\vspace{-4mm}
\clearpage

\appendix

\section{Classes of ImageNet100}

We choose 100 classes from ImageNet ISLVRC2012~\cite{deng2009imagenet} to build ImageNet100 in the paper. The class names of ImageNet100 are: 

\textit{`American robin', `Gila monster', `eastern hog-nosed snake', `garter snake', `green mamba', `European garden spider', `lorikeet', `goose', `rock crab', `fiddler crab', `American lobster', `little blue heron', `American coot', `Chihuahua', `Shih Tzu', `Papillon', `toy terrier', `Treeing Walker Coonhound', `English foxhound', `borzoi', `Saluki', `American Staffordshire Terrier', `Chesapeake Bay Retriever', `Vizsla', `Kuvasz', `Komondor', `Rottweiler', `Dobermann', `Boxer', `Great Dane', `Standard Poodle', `Mexican hairless dog (xoloitzcuintli)', `coyote', `African wild dog', `red fox','tabby cat', `meerkat', `dung beetle', `stick insect', `leafhopper', `hare', `wild boar', `gibbon', `langur', `ambulance', `baluster handrail', `bassinet', `boathouse', `poke bonnet', `bottle cap', `car wheel', `bell or wind chime', `movie theater', `cocktail shaker', `computer keyboard', `Dutch oven', `football helmet', `gas mask or respirator', `hard disk drive', `harmonica', `honeycomb', `clothes iron', `jeans', `lampshade', `laptop computer', `milk can', `mixing bowl', `modem', `moped', `graduation cap', `mousetrap', `obelisk', `park bench', `pedestal', `pickup truck', `pirate ship', `purse', `fishing casting reel', `rocking chair', `rotisserie', `safety pin', `sarong', `balaclava ski mask', `slide rule', `stretcher', `front curtain', `throne', `tile roof', `tripod', `hot tub', `vacuum cleaner', `window screen', `airplane wing', `cabbage', `cauliflower', `pineapple', `carbonara', `chocolate syrup', `gyromitra', `stinkhorn mushroom'}.

\section{More Detailed Results of Experiments}

\subsection{More Detailed Results on Crosss-Datasets Continual Learning}

We also test L2P~\cite{wang2022learning} and S-liPrompts~\cite{wang2022s} on the proposed Cross-Datasets Continual Learning (CDCL) benchmark in this supplementary material. DualPrompt~\cite{wang2022dualprompt} is an updated version of L2P, so we compare with DualPrompt directly, rather than L2P, in the paper. S-liPrompts is a CLIP-based method of domain incremental learning. We treat each dataset as a domain, and test S-liPrompts on the CDCL benchmark.

The complete experimental comparisons on CDCL are reported in Tables~\ref{tab:cifarima_supp}, \ref{tab:inference_supp}, \ref{tab:bothdata_supp}. AttriCLIP obtains the best results in the long-sequence, domain-shift continual learning task. Moreover, it achieves remarkable performance in generalizing to a new dataset and preventing forgetting the previous dataset.

\begin{table}[t]
\centering
\caption{{Accuracy of different methods on CIFAR100. The models are either trained from scratch on CIFAR100 (CIFAR100), or fine-tuned on CIFAR100 after being continually trained from scratch on ImageNet100 (CIFAR100-I2C).}
}
\resizebox{1\columnwidth}{!}{
\begin{tabular}{ccccc}
\toprule[1pt]
Method      & Memory & CIFAR100 & CIFAR100-I2C & FT           \\ \midrule[0.3pt]
\textit{iCaRL}-1       & 2000      & 49.5              & 49.7                   & $+$0.2          \\
\textit{iCaRL}-2       & 2000      & 49.1              & 46.5                   & -2.6          \\
\textit{CoOp}-1        & 1000   & 67.6              &   61.1                  & -6.5         \\
\textit{CoOp}-2        & 1000   &  67.6                 &  59.0                     &     -8.6          \\
\textit{ARI}-1         & 2000   & 80.9              &    74.5                 & -6.4          \\
\textit{ARI}-2         & 2000   &  79.7               &     59.9                  &    -19.8           \\
Continual-CLIP         & 0   & 66.7                &   66.7                   &    0          \\
S-liPrompts & 0      &    {58.9}               &     53.3                     &     -5.6            \\
\textit{L2P}-1 & 0      &    {83.8}               & 78.9                     &   -4.9         \\
\textit{L2P}-2 & 0      &    {80.7}               &     72.4                     &     -8.3            \\
\textit{DualPrompt}-1 & 0      &    \textbf{86.5}               & 80.7                     &   -5.8         \\
\textit{DualPrompt}-2 & 0      &    {84.1}               &     74.7                     &     -9.4            \\

\midrule[0.3pt]
\textbf{AttriCLIP}      & 0      & 81.4     & \textbf{82.3}          & \textbf{$+$0.9} \\ \bottomrule[1pt]
\end{tabular}}
\vspace{-5mm}
\label{tab:cifarima_supp}
\end{table}

\begin{table}[t]
\centering
\caption{{Accuracy of different methods on ImageNet100. The models are either trained from scratch on ImageNet100 (ImageNet100), or fine-tuned on CIFAR100 after being continually trained from scratch on ImageNet100 (ImageNet100-I2C).}
}
\resizebox{1\columnwidth}{!}{
\begin{tabular}{ccccc}
\toprule[1pt]
Method                                              &Memory          & ImageNet100 & ImageNet100-I2C& BT \\ \midrule[0.3pt]
\textit{iCaRL}-1                                                      &2000   & 59.5        & 34.5 & -15.2 
\\
\textit{iCaRL}-2                                                     &2000     & 58.7        & 50.9 &  -7.8                                                                    \\
\textit{CoOp}-1                                                    &1000       & 79.3        & 57.6 &   -21.7                                                                     \\
\textit{CoOp}-2     &1000   & 79.3       & 75.9 &   -3.4                                                                   \\
\textit{ARI}-1      &2000                                                      & 79.3        & 51.2 & -28.1                                                                    \\
\textit{ARI}-2   &2000     & 77.9        & 61.8 &  -16.1                                                                  \\
Continual-CLIP         & 0   &   75.4              &      75.4                &     0         \\
S-liPrompts & 0      &    {61.1}               &     40.9                     &     -20.2            \\
\textit{L2P}-1 & 0      &    {82.7}               & 59.6                     &   -23.1         \\
\textit{L2P}-2 & 0      &    {81.1}               &     74.8                     &     -6.3            \\
\textit{DualPrompt}-1                                             &0    &    \textbf{85.4}            &  63.6    &  -21.8                                                                         \\
\textit{DualPrompt}-2                                             &0       &    81.9           &  77.8    &   -4.1                                                                \\ \midrule[0.3pt]
\textbf{AttriCLIP}                                                 &0         & 83.3        & \textbf{90.3} & \textbf{$+$7.0}                                                               \\\bottomrule[1pt]
\end{tabular}}
\vspace{-2mm}
\label{tab:inference_supp}
\end{table}

\begin{table}[ht]
\centering
\caption{Comparison among different methods on ImageNet100 $+$ CIFAR100 where each model is continually trained on ImageNet100 and CIFAR100 in sequence.}
\resizebox{0.7\columnwidth}{!}{
\begin{tabular}{ccc}
\toprule[1pt]
Method                                                        & Memory &\begin{tabular}[c]{@{}c@{}}CIFAR100$+$\\ ImageNet100\end{tabular} \\ \midrule[0.3pt]
\textit{iCaRL}-1                                                         & 2000          & 30.7                                                                  \\
\textit{iCaRL}-2                                                         & 2000          & 37.6                                                                  \\
\textit{CoOp}-1                                                          & 1000            & 46.6                                                                  \\
\textit{CoOp}-2 & 1000        & 55.4                                                                  \\
\textit{ARI}-1                                                           & 2000             & 32.5                                                                  \\
\textit{ARI}-2  & 2000           & 57.3                                                                  \\
Continual-CLIP         & 0         &  54.9            \\
S-liPrompts         & 0         &  39.6           \\
\textit{L2P}-1                                                &    0                 &      34.9                                                                    \\
\textit{L2P}-2                                                  &    0              &     64.6                                                              \\
\textit{DualPrompt}-1                                                &    0                 &      35.4                                                                    \\
\textit{DualPrompt}-2                                                  &    0              &     67.1                                                              \\ \midrule[0.3pt]
\textbf{AttriCLIP}                                                        & 0          & \textbf{78.3}                                                                  \\\bottomrule[1pt]
\end{tabular}}
\vspace{-5mm}
\label{tab:bothdata_supp}
\end{table}

\section{More Visualization}

\begin{figure}[t]
    \centering
    \includegraphics[width=1\linewidth]{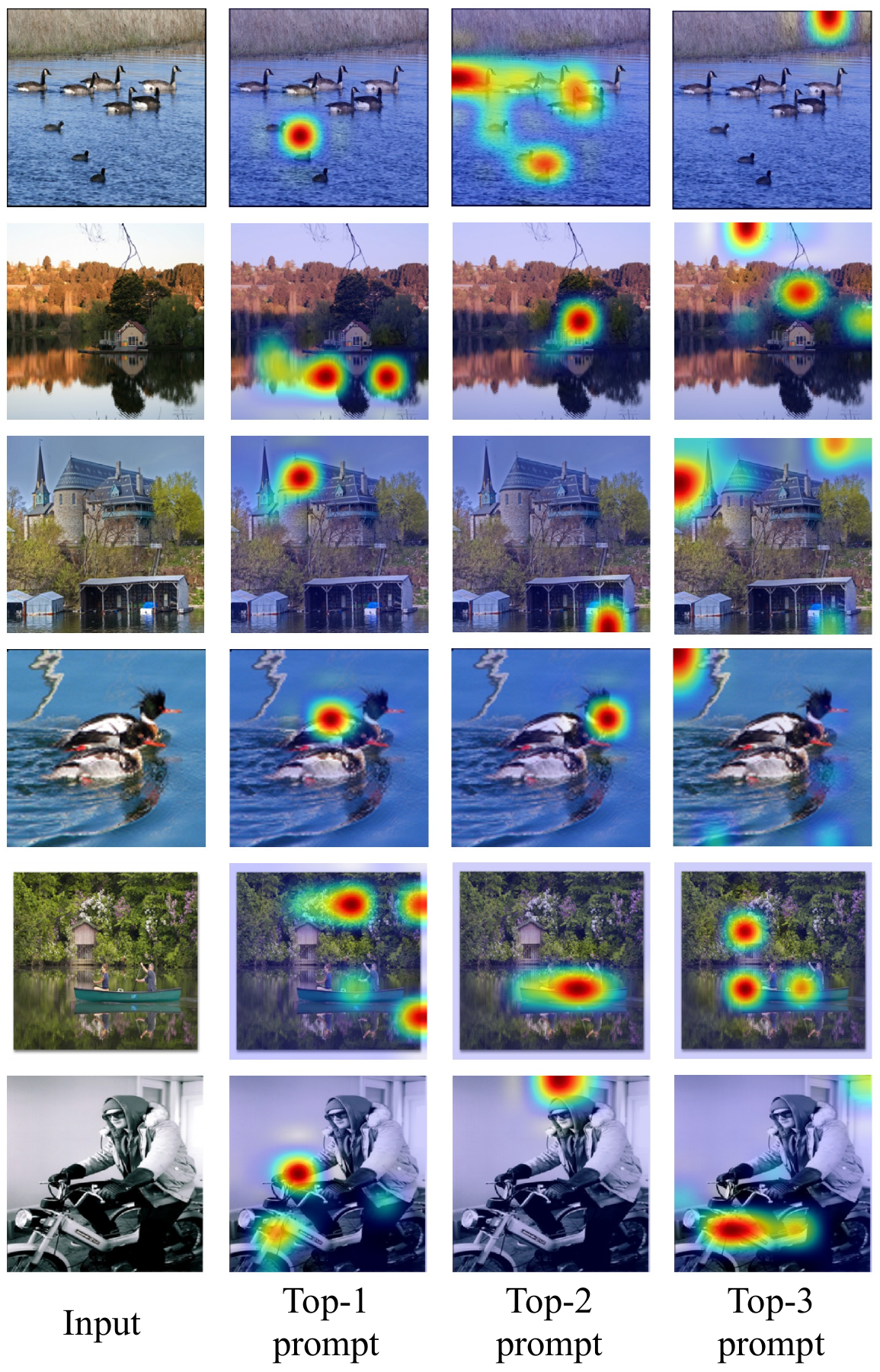}
    \caption{More Visualization of the top three prompts of the same image using Grad-CAM~\cite{selvaraju2017grad}.}
    \label{fig:supp_image}
\end{figure}

\begin{figure}[h]
    \centering
\includegraphics[width=1\linewidth]{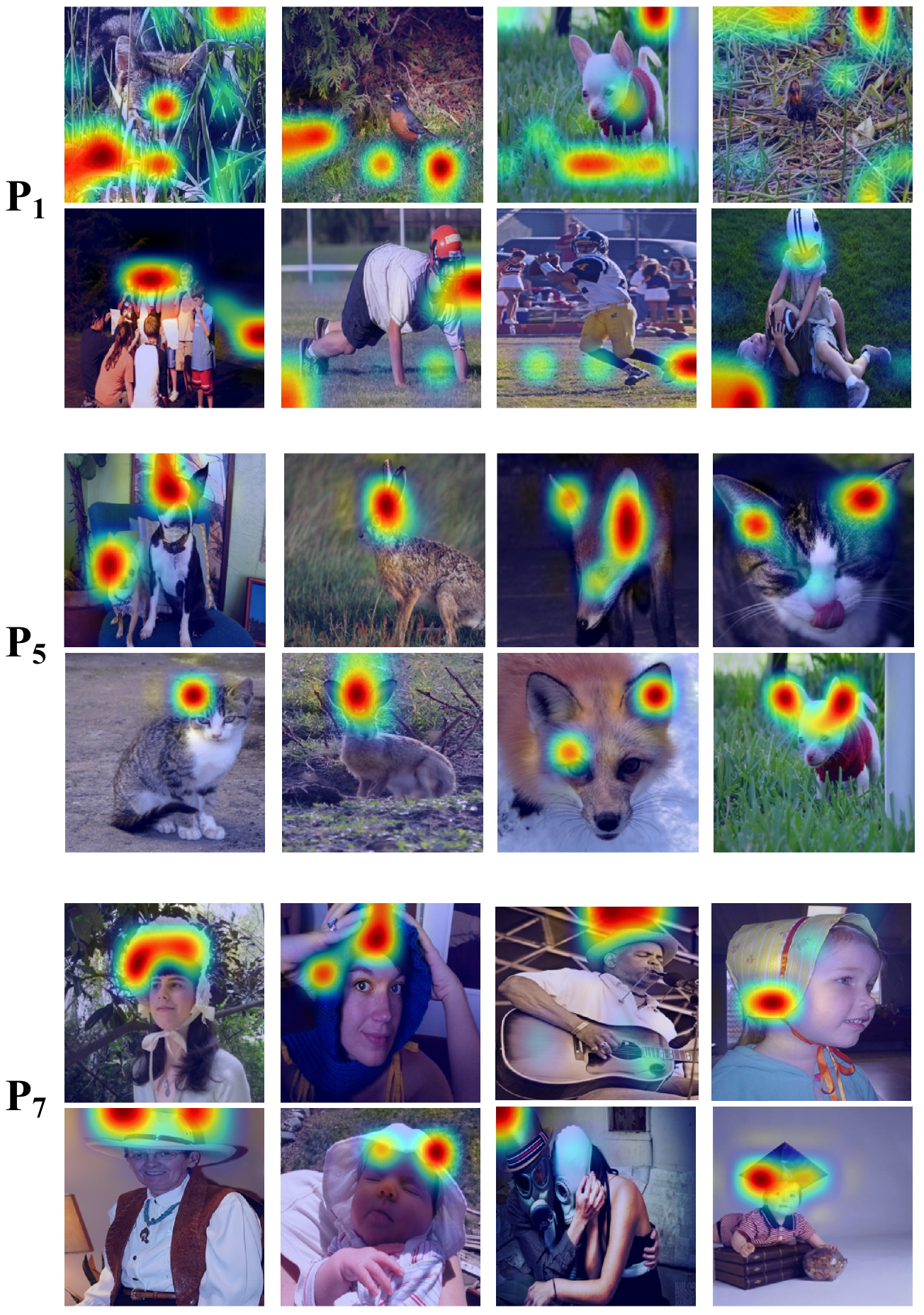}
    \caption{More Visualization of the same prompt on different images using Grad-CAM~\cite{selvaraju2017grad}.}
    \label{fig:supp_prompt}
\end{figure}

In this section, we provide more visualization to verify that different prompts do reflect different image attributes. In Fig.~\ref{fig:supp_image}, the diversity of the learned prompts is verified on more images.
In Fig.~\ref{fig:supp_prompt}, we visualize the contents of more prompts (\emph{e.g.}, $\mathbf{P}_1$, $\mathbf{P}_5$ and $\mathbf{P}_7$) on more images. The newly tested $\mathbf{P}_7$ suggests that it focuses more on the ``hat" content for these images. The combinations of $\mathbf{P}_1$, $\mathbf{P}_2$, ..., $\mathbf{P}_N$ can represent different contents of images after training.

\section{Limitation and Ethic Impact}
Although AttriCLIP is a non-incremental learner, the size of the attribute bank is related to the length of the task sequence. For long sequence tasks, a larger attribute bank is required to cover the necessary attributes in the tasks for learning. In the future, we will test AttriCLIP on more datasets and on longer-sequence CDCL benchmarks.
This work does not have obvious ethic impacts.

\clearpage

{\small
\bibliographystyle{ieee_fullname}
\bibliography{egbib}
}

\end{document}